%% file: main.tex
\documentclass[a4paper,fleqn]{cas-dc}

\usepackage{amssymb}
\usepackage{amsmath}
\usepackage{url}
\usepackage{xcolor}
\usepackage[normalem]{ulem}

\newcommand{\textcode}[1]{\texttt{#1}}

\begin{document}



\let\WriteBookmarks\relax
\def\floatpagepagefraction{1}
\def\textpagefraction{.001}
\shorttitle{Logic Augmented Generation}
\shortauthors{A. Gangemi and A.G. Nuzzolese}

\title [mode = title]{Logic Augmented Generation}                      


\author[AG,AGN]{Aldo Gangemi}
    [orcid=0000-0001-5568-2684]
    \fnref{fn1}
    \ead{aldo.gangemi@cnr.it}
    \ead[url]{https://istc.cnr.it/it/people/aldo-gangemi}

\author[AGN]{Andrea Giovanni Nuzzolese}
    [orcid=0000-0003-2928-9496]
    \fnref{fn1}
    \ead{andreagiovanni.nuzzolese@cnr.it}
    \ead[url]{https://istc.cnr.it/it/people/andrea-nuzzolese}

\affiliation[AG]{organization={University of Bologna},
            city={Bologna},
            country={Italy}}
\affiliation[AGN]{organization={CNR - Institute of Cognitive Sciences and Technologies},
            city={Bologna},
            country={Italy}}

\fntext[fn1]{This author contributed equally to this work.}

\begin{abstract}
Semantic Knowledge Graphs (SKG) face challenges with scalability, flexibility, contextual understanding, and handling unstructured or ambiguous information. However, they offer formal and structured knowledge enabling highly interpretable and reliable results by means of reasoning and querying. Large Language Models (LLMs) may overcome those limitations, making them suitable in open-ended tasks and unstructured environments. Nevertheless, LLMs are hardly interpretable and often unreliable. To take the best out of LLMs and SKGs, we envision Logic Augmented Generation (LAG) to combine the benefits of the two worlds. LAG uses LLMs as Reactive Continuous Knowledge Graphs that can generate potentially infinite relations and tacit knowledge on-demand. LAG uses SKGs to inject a discrete heuristic dimension with clear logical and factual boundaries. We exemplify LAG in two tasks of collective intelligence, i.e., medical diagnostics and climate projections. Understanding the properties and limitations of LAG, which are still mostly unknown, is of utmost importance for enabling a variety of tasks involving tacit knowledge in order to provide interpretable and effective results.
\end{abstract}


\maketitle



\section{Introduction}
\label{sec:intro}
\input{sections/intro}

\section{Use case and problem framing}
\label{sec:problem}
\input{sections/problem}

\section{LLMs as Reactive Continuous Knowledge Graphs with logic boundaries}
\label{sec:rckg}
\input{sections/rckg}

\section{LAG: SKG+RCKG}
\label{sec:lag}
\input{sections/lag}

\section{Conclusions and future work}
\label{sec:conclusions}
\input{sections/conclusions}

\section*{Acknowledgements}
This work was supported by the European Union’s Horizon Europe research and innovation programme within the context of the project HACID (Hybrid Human Artificial Collective Intelligence in Open-Ended Domains, grant agreement No 101070588), and FAIR (Future Artificial Intelligence Research PNRR project, code PE00000013 CUP 53C22003630006)

\bibliographystyle{splncs04}
\bibliography{references}

\end{document}

%% file: sections/intro.tex
Although there is no general consensus on the exact definition of Knowledge Graph (KG) in literature~\cite{Ji2021}, we follow ~\cite{Hogan2021}, and define KGs as linked data (typically expressed in the Web Ontology Language - OWL and the Resource Description Framework - RDF triples) including both schema  and assertional axioms. We refer to these KGs as Semantic Knowledge Graphs (SKGs), because of their rigorous model-theoretical semantics.
SKGs provide structured, interpretable knowledge with logical querying and reasoning capabilities, making them valuable tools for modelling rich, linked datasets in any domain. 
However, their limitations in scalability, flexibility, and the integration of non-standard input restrict their application in open-ended tasks requiring complex  extraction of perspectival and incomplete  knowledge. {\em Collective intelligence}~\cite{Leimeister2010}, as used in medical diagnostics and climate services is such a case. Additionally, SKGs: (i) are typically static and cannot natively represent dynamic, evolving knowledge (e.g. they struggle with temporal reasoning), making it difficult to handle changes over time, such as tracking historical trends or future predictions; (ii) have limited ability to represent uncertainty or probabilistic relationships (Fuzzy OWL is an example~\cite{Stoilos2005}), which are often crucial in real-world scenarios and open-ended tasks; (iii) do not provide built-in mechanics to handle conflicting information from multiple sources\footnote{This can lead to inconsistencies and difficulty when reconciling diverse expert contributions or contradictory data.}; (iv) are short of implicit or common-sense knowledge, which humans rely on extensively, thus reducing their effectiveness in tasks requiring contextual understanding or implicit knowledge extraction; (v) excel at representing explicit knowledge but are inadequate for extracting or encoding tacit knowledge, which is often crucial for decision-making and problem-solving in complex domains.
To address these challenges, we introduce {\em Logic Augmented Generation} (LAG), a novel paradigm that integrates SKGs with Large Language Models (LLMs) to leverage the strengths of both approaches. LAG conceptualises LLMs as potential {\em Reactive Continuous Knowledge Graphs} (RCKGs), which can dynamically adapt to diverse inputs by extending and contextualising SKGs that work as base models. SKGs ensure logical consistency, enforce factual boundaries, and foster interoperability, while LLMs process unstructured data and provide contextual insights and tacit knowledge on demand. This dual approach enhances interpretability and reliability, mitigating the lack of truth-theoretic semantics in LLM's output. We expect that LAG methods enable effective collaboration among experts and support the co-creation of actionable insights in complex, evolving contexts such as climate adaptation and collective medical decision-making. 

%% file: sections/problem.tex
Semantic Knowledge Graphs (SKGs) provide structured, interpretable knowledge and support precise, logical querying. They have been used for modelling rich linked datasets in a variety of domains, such as health, cultural heritage, social and climate sciences, etc. In the context of the HACID\footnote{The HACID acronym stands for Hybrid Human Artificial Collective Intelligence in Open-Ended Decision Making project.} project~\cite{kurvers2023}, we designed and modelled an SKG to support experts in open-ended diagnostic tasks with a {\em collective intelligence}~\cite{Leimeister2010} approach in two different domains, i.e., medical diagnostics and climate services. Collective intelligence is the shared, tacit, or group intelligence that emerges from the collaboration, collective efforts, and interactions of a group of individuals or entities. This concept is rooted in the idea that a group whose members work together effectively can achieve insights, solve problems, or make decisions that are beyond the ability of a single individual. 

We use the following simple medical case, referred to as CASE, as an example to clarify the problem.

\vspace{0.2cm}
{\em``A 38-year-old male presents with fever and a dry cough that has persisted for the last 4 days. He has no significant medical history, although he recently returned from a business trip.''}
\vspace{0.2cm}

In the medical case above, the business trip is the potential cause of the patient's medical condition. However, the causality relation is inferred by recognising the potential tacit connection between the business trip and the illness (such as travel-related infections) without it being directly stated or formally encoded in the information provided. It also includes the ability of physicians to identify and weigh subtle clues or patterns that may not be immediately obvious from the presented data.

Consequently, the objective is to use an SKG to smoothen the collaboration and communication among experts by harmonising the diversity of thought in a context in which no single individual has all the answers, but the group as a whole possesses the knowledge needed to solve a problem. 
In medical diagnostics, the ultimate goal is to identify the correct diagnosis for a patient, thereby reducing errors, preserving life, and minimising the costs associated with incorrect treatments. 

Climate services urgently require systems that leverage collective intelligence to navigate and integrate vast, diverse datasets. By fostering collaboration among scientists, stakeholders, and AI-driven tools, challenges in the interpretability of results and the handling of unstructured data can be addressed more effectively.

To address those objectives, our SKG integrates different existing knowledge bases and provides a coherent conceptualisation formalised in a rich modular ontology network designed by reusing Ontology Design Patterns~\cite{Gangemi2009} (ODPs) and leveraging DOLCE-Zero~\cite{paulheim2015serving} as a top-level ontology. For medical diagnostics, we have integrated data gathered from SNOMED-CT, ICD-10, and Wikidata. For climate services, we have integrated data from the Coupled Model Intercomparison Project\footnote{\url{https://www.wcrp-climate.org/wgcm-cmip}.} (CMIP)~\cite{Eyring2016}. 

Nevertheless, the effectiveness of an SKG solely based on symbolic knowledge representation and reasoning remains limited when we need to support open-ended tasks with expansive problem spaces, such as collective intelligence-based medical diagnostics, or climate services. 
This limitation arises from the challenges of harmonizing and integrating standardised knowledge in the reference SKG with non-standard, natural language inputs from multiple experts. Firstly, the different inputs provided by experts might express different viewpoints in different languages and factual granularities. Secondly, tacit and domain-specific background knowledge might be required in order to make sense of those inputs, and making a superior collective knowledge emerge. 
Moreover, SKGs struggle with scalability, flexibility, evolvability, and the integration of complex, context-rich information. For example, climate data include real-time environmental observations, policy impacts, and social adaptation strategies. 

Large Language Models (LLMs), in contrast, can be employed to manage unstructured data and to generate novel insights on demand, making them well-suited for open-ended tasks. Yet, their lack of interpretability and reliability hinders their application in fields where precise and accountable outputs are critical. 

%% file: sections/rckg.tex
To address this gap, we use LLMs as potential {\em Reactive Continuous Knowledge Graphs} (RCKGs). 
RCKGs are SKGs extracted from multimodal signal (e.g. text, speech, pictures, sensory data, etc.), using (typically or mostly) continuous vector spaces, such as generative pre-trained transformers.
RCKG extraction involves a three-step process, i.e. (i) mapping multimodal signal to natural language, (ii) converting natural language into an SKG, and (iii) extending the SKG with tacit knowledge based on multiple heuristics. Each transformation in this pipeline plays a crucial role in enabling the RCKG's ability to generate dynamic, context-sensitive knowledge representations. The first transformation, mapping signal to natural language, is characterised as {\em supramodal} \cite{FairhallCaramazza2013}\cite{Binder2016} because natural language inherently integrates and synthesises information from multiple sensory modalities (e.g., visual, auditory, tactile inputs) by representing objects and concepts that are invariant across different sensory modalities. For instance, a textual description of an image consolidates some visual details, or a written summary of a sound encapsulates certain auditory characteristics. Natural language serves as a unified, modality-transcending medium that bridges sensory-specific data and symbolic representation. This supramodal nature makes natural language an ideal intermediary for integrating multimodal signals into a coherent, human-readable format that captures both explicit information and implicit nuances, which are more or less easily reconstructed by human interpreters. The second transformation, converting natural language to an SKG, is defined as {\em amodal} \cite{Binder2016} because knowledge graphs abstract knowledge from its sensory or linguistic origins. In this step, semantic relationships and entities are encoded in a structured, modality-neutral format that is fully machine-interpretable. This abstraction ensures that the resulting representation is independent of the modalities through which the original information was perceived or described. The amodal KG thus becomes a foundation for logical reasoning, enabling precise querying, inference, and integration with external knowledge sources, such as SKGs. The third transformation, extending the SKG with implicit/tacit knowledge, aims at recovering the tacit dimension of  natural language as used in context, by leveraging reactive natural language models such as generative LLMs as providers. This extension results in Reactive Continuous Knowledge Graphs (RCKGs). The third transformation again exploits the supramodal nature of language to generate amodal knowledge.

The relevance of this supramodal-to-amodal pipeline in RCKG lies in its ability to leverage the strengths of both transformations. By systematically aligning supramodal and amodal transformations, a RCKG provides a scalable and versatile framework for synthesising, abstracting, and reasoning with multimodal inputs, thereby advancing the potential of neuro-symbolic systems in dynamic and open-ended tasks.

An SKG can be generated from text in multiple ways. State-of-the-art SKG extraction from text assumes \emph{situations} as occurrences of conceptual \emph{frames}, following the knowledge extraction paradigm introduced by the FRED machine reader~\cite{Gangemi2017}, which extracts OWL SKGs based on Framester Semantics~\cite{Gangemi2016}, and is also used~\cite{gangemi2023taf} to formalise Abstract Meaning Representation~\cite{Banarescu2013} (AMR) graphs.
Once extracted, an SKG is passed (within an augmented prompt) to an LLM as a source of extended meaning, to contextualise and complete an SKG with implicit and local knowledge. In this sense, a LLM is used as a reactive AI, which generates a piece of extended knowledge $K_{ext}$ when stimulated by an input piece of knowledge $K_{inp}$. 

A first implementation of LAG as SKG-augmented RCKG generation is demoed as a multimodal knowledge extractor \footnote{https://arco.istc.cnr.it/itaf/} \cite{itaf2024}, used to generate the example graph in Figure \ref{fig:example}.

While the notions of reactivity and continuity in Knowledge Graphs (KGs) have established meanings in the literature, Reactive Continuous Knowledge Graphs (RCKGs) add complementary interpretations.

The reactivity of a Knowledge Graph (KG) has typically been associated with notions of dynamism and temporal awareness. A notable example is the work in~\cite{Krause2022}, which explores the extension of KGs to accommodate temporal and dynamic aspects. Its authors focus on explicitly encoding time-aware data through timestamps or mappings across time points, thereby preserving historical and future contexts. More broadly, Reactive Continuous Knowledge Graphs (RCKGs) integrate tacit knowledge and dynamically adapt it using LLMs. RCKGs capture reasoning and contextual nuances implicit in data by generating new knowledge adaptively, blending probabilistic and logical reasoning in a manner that extends beyond static temporal annotations.

In addition to our definition of reactive and (mathematically) continuous KGs, Continuous Knowledge Graphs have been introduced recently, e.g.  in~\cite{Lairgi2024} as evolutionary dynamic graphs, featuring incremental knowledge updates, schema adaptation, conflict resolution, source tracking and verification, and temporal awareness. The authors propose to reduce extraction to the normalization and manipulation of elementary propositions, but recognize that this is a very challenging task. As said above, logic augmented generation is better suited to represent \emph{situations} as occurrences of conceptual \emph{frames} in a frame semantics-compatible logic like OWL2. 

The notion of continuity proposed by~\cite{Lairgi2024} focuses on the process of updating  dynamic and adaptable knowledge representation systems that can evolve and integrate new information systematically over time. The authors in~\cite{Lairgi2024} discuss the well-known task of continuously evolving graphs over time using available information. 
In addition, RCKGs are continuous firstly because LLMs are reactive continuous models: they are mostly trained on a finite and discrete corpus of data, which forms the basis of their learned knowledge; however, due to their ability to generalise from training data, they can generate a potentially infinite number of outputs by combining concepts, patterns and relationships in novel ways. This generative capability is what allows LLMs to learn and generate responses based on context-sensitive input data (such as text) and to continuously adapt to new contexts or queries during in-context learning. Hence, by means of prompting, their internal representation of knowledge adapts dynamically as more data is processed, allowing them to dynamically adjust and generate relevant responses based on the latest input. 
Accordingly, if properly prompted to generate a KG, a LLM can generate potentially infinite triples by means of in-context learning. This means, for example, that LLMs do not require retraining, adjustments or evolution to their underlying data to generate new knowledge. More specifically, this characteristic refers specifically to the ability of LLMs to contextually interpret and generate knowledge based on input prompts, rather than performing fundamental updates to their internal weights or training data. This adaptation is achieved through in-context learning, where the model dynamically adjusts its responses based on the structure and content of the input without altering its pre-trained parameters. LLMs with fixed weights rely on their generative capabilities, drawing from a latent knowledge space encoded during training. While this space is finite and static, the models are capable of producing effectively infinite outputs by combining and contextualising learned patterns in novel ways. This makes them highly flexible for generating new knowledge dynamically, in response to specific prompts, without requiring retraining or modifications to the underlying knowledge corpus. In contrast, approaches like fine-tuning or continual learning involve fundamental updates to the model's weights, enabling the incorporation of new knowledge at the cost of additional computational overhead and potential domain-specific constraints. By focusing on prompting and in-context learning, LAG leverages on fixed-weight LLMs to serve as RCKGs, adapting dynamically to new inputs without requiring structural adjustments to the model.

%% file: sections/lag.tex
\begin{figure*}[ht!]
  \centering
  \includegraphics[width=0.9\textwidth,height=0.9\textheight, keepaspectratio]{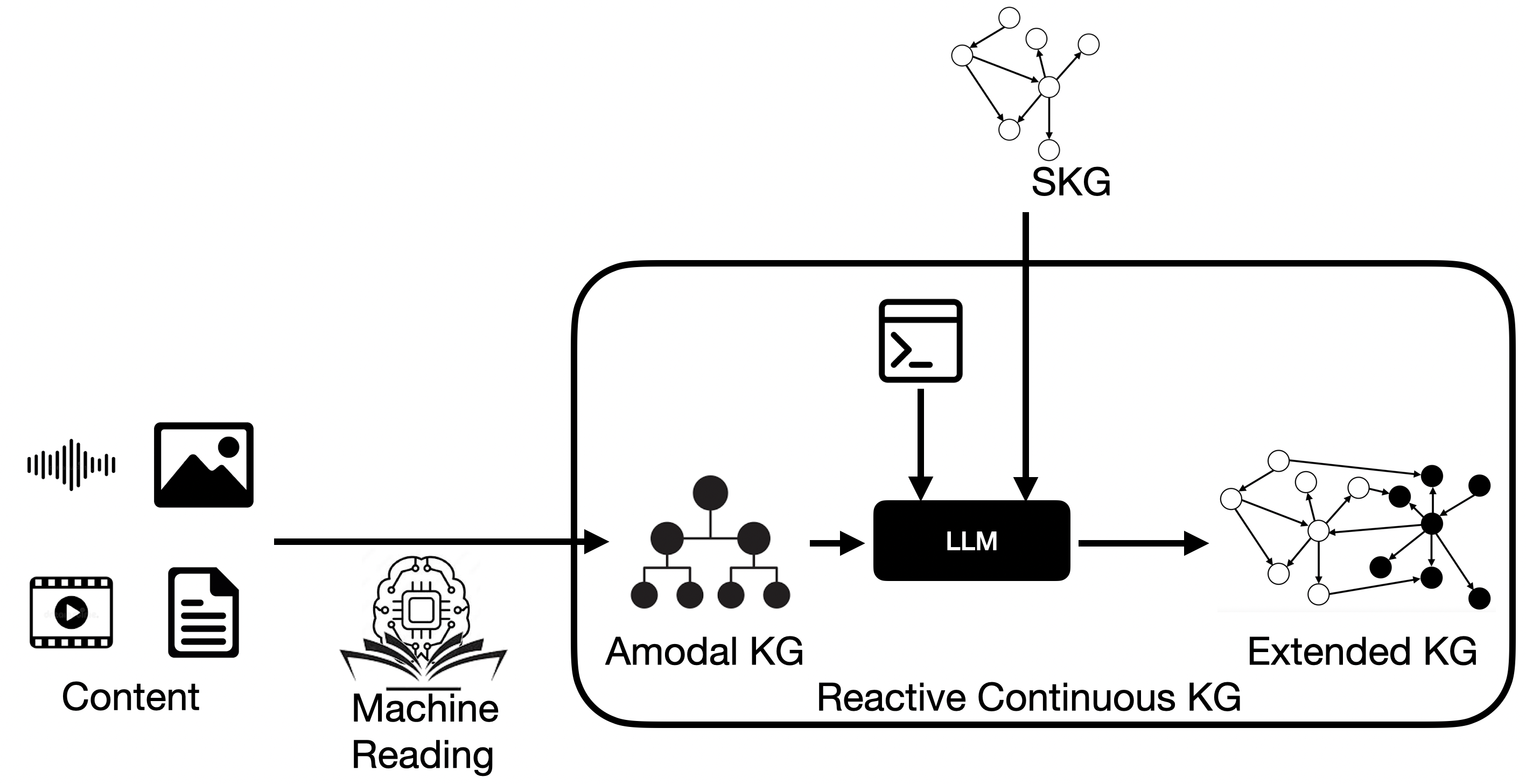}
  \caption{The architecture of LAG.}
  \label{fig:rckg}
\end{figure*}

RCKGs are an example of a neuro-symbolic approach that generates a much richer and deeper knowledge graph than the basic meaning of the original text. RCKGs extract tacit knowledge by capturing insights, reasoning, or behaviours that resemble the informal, experiential, and often unspoken knowledge humans acquire through experience and practice \cite{polanyi2009tacit}. While tacit knowledge is difficult to articulate or codify, it plays a crucial role in problem-solving and decision-making, making it particularly valuable for collective intelligence tasks that require collaboration, intuition, creativity, and experience. 
In medical diagnostics, tacit knowledge could involve a clinician’s ability to identify subtle patterns that suggest a rare disease based on prior experience or an intuitive sense of urgency derived from non-verbal cues of a patient. For example, while symptoms such as fever and cough can be codified in structured medical records, a doctor’s recognition that recent travel increases the likelihood of a tropical infection involves tacit knowledge. Similarly, in climate services, tacit knowledge includes understanding the implicit risks associated with certain weather patterns. For example, a meteorologist might intuitively anticipate that an unusual sea surface temperature anomaly could lead to intensified tropical storms, even before formal models confirm it. Similarly, local knowledge about how specific communities respond to extreme weather events is often unspoken, but vital for effective planning and adaptation strategies.
Verifying and interpreting tacit knowledge is challenging, particularly in terms of reliability and ethical considerations, as it is not derived from human personal experience, validated sensors, or deductive inferences, but from patterns in data.

In the case of LLMs, tacit knowledge is implicitly encoded in the vast training data they process. LLMs do not {\em understand} tacit knowledge as humans do, but capture and generalise patterns, correlations, and relationships present in the data~\cite{Chang2024}. These models are trained on diverse multimodal corpora, which often include fragments of tacit knowledge embedded in human language, such as metaphors, cultural nuances, or context-specific inferences. This training phase does not explicitly codify tacit knowledge but instead creates a latent space where such patterns are represented implicitly as probabilistic associations across continuous vector spaces. Hence, we assume that the generative capabilities of LLMs allow them be used as RCKG to make tacit knowledge more explicit when prompted effectively. Hence, by leveraging the ability of LLMs to synthesise patterns from training data, RCKG can generate representations (i.e. triples) that resemble the tacit insights humans infer from their experiences. For example, when presented with a prompt that implicitly contains causal relationships or contextual nuances, the model can produce explicit knowledge triples or statements that align with those relationships. This process can be understood as a {\em ``reactive codification''}, where tacit knowledge is externalised as explicit, structured outputs in response to specific inputs. In a nutshell, LAG exploits the duality between implicit tacit knowledge (i.e. the one generated during training) and explicit tacit knowledge (i.e. the one externalised in the form of structured knowledge graphs through generation and prompting). Accordingly, LAG use RCKGs to surface tacit knowledge that would otherwise remain latent in the training data. The process hinges on prompt engineering and in-context learning, which guide the model to retrieve, synthesise, and represent tacit patterns relevant to a specific query or context.
Nevertheless, a semantics for RCKG tacit knowledge requires addressing the nature of reactive inferences in a continuous space with discrete input and output that can be context-sensitive.

This puts RCKG apart from most logics. For example, RCKG semantics is {\em plausibility-preserving} rather than {\em truth-preserving}. Its operators for entailment, necessity, possibility, etc. would be interpreted differently from traditional truth-theoretic ones, and the interaction between classical truth-preserving and plausibility-preserving axioms needs to be negotiated. E.g., plausibility does not guarantee truth, while truth guarantees plausibility, leading to potential invalidity of mixed transitivity chains. Also, tacitness is typically non-monotonic and context-dependent. 
Furthermore, multiple tacit operators emerge which can vary in their reasoning behaviour: social implicatures can have different logical properties (group relativity, role sensitivity, status dependency, intensity, valence, gradability, source dependency, etc.) from affective evocations, evaluative judgments, causal consequences, metaphorical blending, etc. However, being able to represent such a variety of tacit knowledge by using a unique continuous model, and forcing it to adhere to the valid logical axioms of an amodal KG is a big step forward to use logic in practical, large scale, and flexible contexts that depend on sophisticated implicit knowledge.

\begin{figure*}[ht!]
  \centering
  \includegraphics[width=0.9\textwidth,height=0.9\textheight, keepaspectratio]{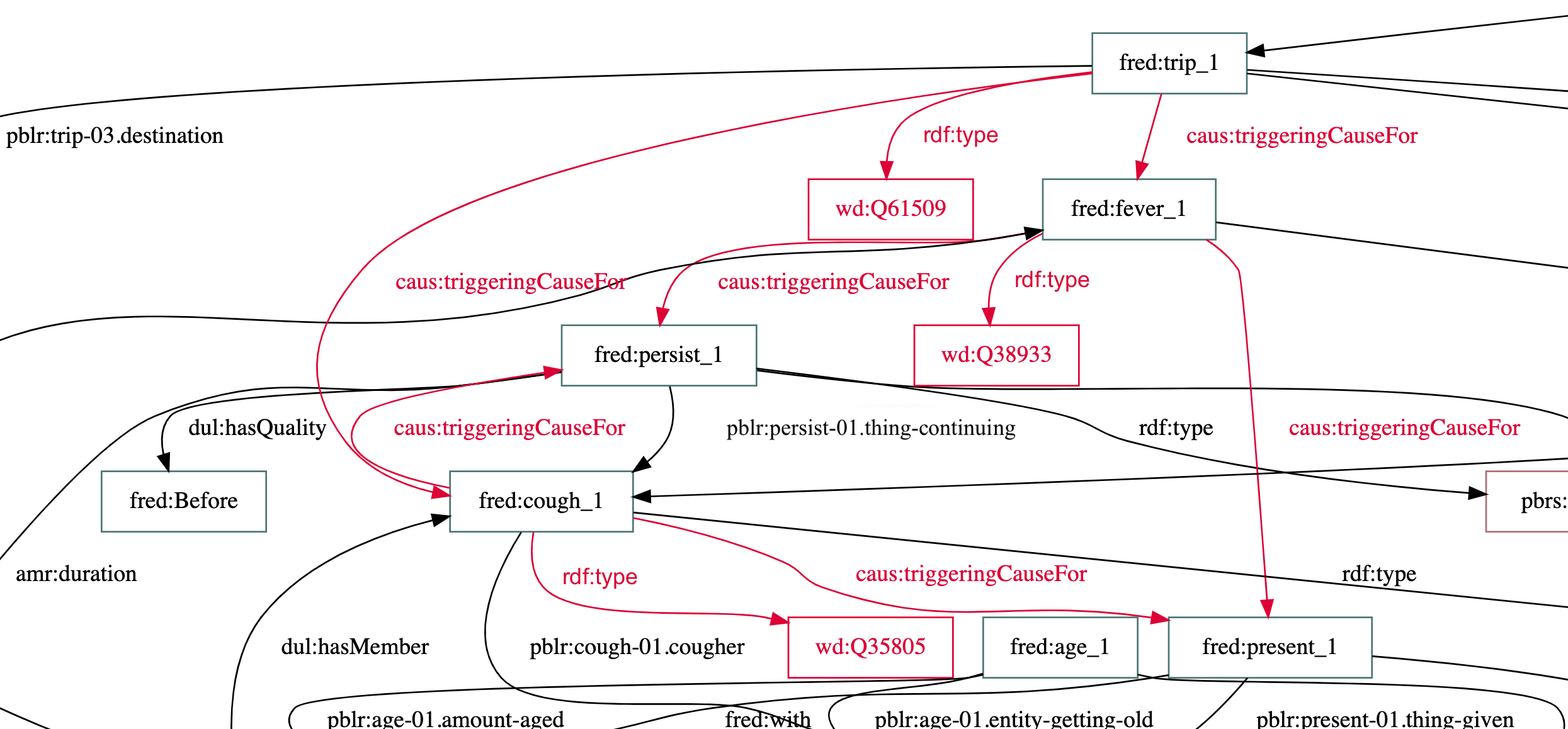}
  \caption{Example of possible extended KG generated by LAG.}
  \label{fig:example}
\end{figure*}

Constraining RCKGs to generate knowledge within precise boundaries poses significant challenges. However, we deem the effort with formalising tacit knowledge valuable. This involves limiting the continuous geometric space to a discrete dimension, which allows, for instance, to get only the relevant knowledge with respect to a given objective (e.g., generating climate projections that harmonise experts' opinions and target expected environmental scenarios). 
To address these issues we started formalising and implementing a novel architecture called Logic Augmented Generation (LAG). LAG integrates RCKGs with Semantic KGs (SKGs) that serve as a discrete, heuristic layer for in-context learning within RCKG's continuous knowledge space, enforcing hybrid logical consistency, factual boundaries, and fostering interoperability with existing SKGs in the Semantic Web.
Figure~\ref{fig:rckg} exemplifies the architecture of a LAG that is composed of: (i) an amodal SKG automatically gathered from text using FRED, representing user queries; (ii) an LLM that is prompted with the amodal SKG, additional prompt heuristics that can be used to refine the context incrementally, and the SKG scoping the desired semantics and providing existing factual knowledge; (iii) an extended SKG produced by LAG, which contains newly generated triples that comply with the SKG, and addresses the desired user queries.
An SKG is enriched as a RCKG via in-context learning that requires adequate prompt engineering strategies. The study of effective prompt engineering strategies to inject the logical and factual knowledge of an SKG into a RCKG is another challenge that LAG must address. Evolutions of Chain-of-Thought, such as Metacognitive Prompting~\cite{Wang2023}, can be beneficial to LAG, as shown by preliminary results presented in~\cite{Lippolis2024}. 
LAG differs from existing paradigms available in literature such as Retrieval-Augmented Generation (RAG)~\cite{Lewis2020} and fine-tuning, addressing critical gaps to enable dynamic and context-sensitive knowledge generation. In fact, RAG enhances LLMs by retrieving relevant information from a fixed corpus to ground responses in explicit factual knowledge. However, it is limited to static corpora and lacks mechanisms to generate or adapt knowledge dynamically. In contrast, LAG leverages RCKGs, which enable dynamic knowledge synthesis by reacting to input prompts, and integrates them with SKGs to ensure logical consistency and factual alignment. Unlike fine-tuning, which requires modifying model weights and retraining to incorporate new knowledge, LAG operates through in-context learning, injecting SKGs as structured prompts into RCKGs. This approach preserves the adaptability and domain neutrality of LLMs while enabling real-time, context-aware knowledge generation. Furthermore, while fine-tuning risks inconsistencies across domains and RAG relies on static retrieval, LAG supports reasoning over evolving knowledge spaces and accommodates tacit knowledge, probabilistic inferences, and hybrid logical reasoning. Additionally, by using a neuro-symbolic approach, LAG facilitates the integration of structured knowledge with generative reasoning, allowing it to address challenges such as handling dynamic and conflicting information. This positions LAG as a complementary and extended paradigm, particularly suited for applications requiring both knowledge generation and reconciliation, such as collective intelligence and collaborative decision-making.

\begin{figure*}[bt!]
  \centering
  \includegraphics[width=0.9\textwidth,height=0.9\textheight, keepaspectratio]{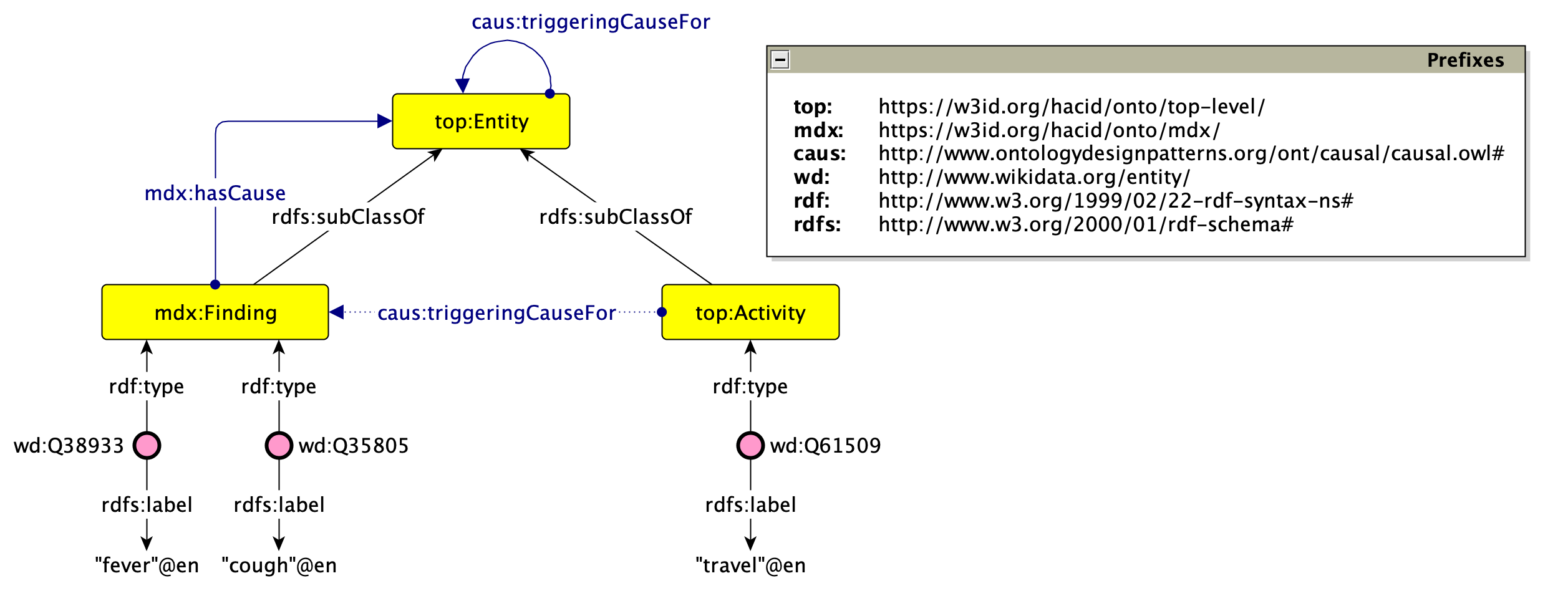}
  \caption{A part of the SKG used as input of LAG to demonstrate  our example (i.e. CASE).}
  \label{fig:skg_example}
\end{figure*}

An example that explains how LAG supports collective intelligence for medical diagnostics is based on the simple medical case we presented in Section~\ref{sec:problem}, i.e. CASE. 
A possible extended KG generated by LAG for CASE is presented in Figure~\ref{fig:example}. In this figure, the amodal KG generated through machine reading is represented by the black nodes and arrows, while the red ones are added to the extended KG by LAG. A part of the SKG used as input of LAG is depicted as a Grafoo\footnote{\url{https://essepuntato.it/graffoo/}}~\cite{Falco2014} diagram in Figure~\ref{fig:skg_example} that shows some core concepts with their semantics. Those concepts are: (i) \textcode{top:Entity}, which is the top level class in our ontology network; (ii) \textcode{mdx:Finding}, which is the class for representing any possible finding associated with medical relevance; and (iii) \textcode{top:Activity}, which represents any action or task planned or executed by an agent. The ontology network used for the medical case is described in a project deliverable~\cite{Nuzzolese2024} and can be retrieved on GitHub\footnote{\url{https://github.com/hacid-project/knowledge-graph/blob/main/ontologies/medical-dx/mdx.owl}}. Additionally, Wikidata entities are used to populate the extensional layer of the SKG. Specifically, the entities \textcode{wd:Q61509}, \textcode{wd:Q38933}, and \textcode{wd:Q35805} are sourced from Wikidata and represent the concepts {\em travel}, {\em fever}, and {\em cough}, respectively. The object properties \textcode{caus:triggeringCauseFor} and \textcode{mdx:hasCause} represent general and specific causal relationships, respectively. 
Indeed, the extended KG captures causal consequences generated as tacit knowledge by the RCKG component of LAG, i.e., the predicates \textcode{caus:triggeringCauseFor}, from the presented medical case.
Then, the reconciliation with Wikidata performed by LAG involves the following steps: (i) entity matching and selection in which relevant entities from Wikidata are identified through entity linking techniques. For instance, terms like ``fever'' or ``cough'' in the medical use case are mapped to their corresponding Wikidata entries (e.g., \textcode{wd:Q38933} for fever), thus ensuring semantic alignment between the input data and the SKG; (ii) contextual-relevant knowledge gathering in which a subset of Wikidata is selected based on the contextual relevance of entities to the task or domain. For example, in the medical diagnostics use case, entities related to diseases, symptoms, and potential causes (such as travel history) are prioritised. This subset can be extracted, for example, using SPARQL queries that traverse relationships and filter entities based on predefined criteria (e.g., medical conditions with known causes or symptoms); (iii) knowledge harmonisation that allows the extracted subset of Wikidata to be incorporated into the extended KG. 
In the example in Figure~\ref{fig:example} the extended KG makes formally emerge that a trip is the potential triggering cause for both fever and cough, a cause that was only tacitly assumed in the case description. Hence, experts might rely on the extended KG in order to provide their opinions that can be aggregated to the extended KG following the same process. This approach has the potential to foster collaboration and improve decision-making in domains such as medical diagnostics and climate services, where the integration of multiple experts’ views, encompassing both tacit and formal knowledge, is crucial.
It is worth noting that implementing LAG requires facing several challenges. First, SKG might contain broad, general-purpose knowledge, which requires filtering irrelevant or noisy data. With respect to this point, frame-based approaches have been demonstrated to be effective to get meaningful knowledge boundaries around broad data. An example is provided in~\cite{Nuzzolese2017}, which uses knowledge patterns, a specific type of frame, to filter relevant contextual knowledge from DBpedia based on user queries. Another challenge to address is the reconciliation between the amodal KG and the SKG (e.g., Wikidata) to preserve the structure and specific schema of the latter by consistently mapping relationships and types. Finally, efficiently navigating and querying a large SKG (e.g., Wikidata) requires optimised query strategies and pre-computed indexes.

%% file: sections/conclusions.tex
In this work we introduce Logic Augmented Generation (LAG), a novel solution that integrates Semantic Knowledge Graphs (SKGs) with Reactive Continuous Knowledge Graphs (RCKGs) to enforce logical consistency and blend symbolic and continuous reasoning. RCKGs are products of a neuro-symbolic method that uses Large Language Models (LLMs) to extract and represent knowledge, incorporating tacit insights and contextual nuances. This hybrid model enables context-aware knowledge generation and addresses challenges like formalising tacit knowledge, enhancing prompt engineering, and aligning plausibility- and truth-preserving logic. The potentiality of LAG is illustrated by using collective intelligence as a case study. Future work will study SKG-RCKG integration, advance prompt strategies, and expand operators for representing tacit semantics.